\documentclass[11pt, letterpaper]{arxiv}
\usepackage[all]{hypcap}
\usepackage[comma,numbers,sort,compress]{natbib}

\usepackage{hyperref}[citecolor=magenta]
\usepackage[capitalize,noabbrev]{cleveref}
\usepackage{xurl}

\hypersetup{
    colorlinks = true,
    citecolor = {magenta},
}

\usepackage{microtype}
\usepackage{subcaption}
\usepackage{booktabs} %

\usepackage[noend]{algpseudocode}
\usepackage{algorithm}
\usepackage{mathrsfs}
\usepackage{setspace}

\usepackage{amsmath}
\usepackage{amssymb}
\usepackage{mathtools}
\usepackage{amsthm}
\usepackage[inkscapelatex=false]{svg}

\setboolean{logo}{true}

\usepackage[most,skins,theorems]{tcolorbox}
\tcbset{
  aibox/.style={
    width=\linewidth,
    top=7pt,
    bottom=2pt,
    colback=blue!6!white,
    colframe=black,
    colbacktitle=black,
    enhanced,
    center,
    attach boxed title to top left={yshift=-0.1in,xshift=0.15in},
    boxed title style={boxrule=0pt,colframe=white,},
  }
}
\newtcolorbox{AIbox}[2][]{aibox,title=#2,#1}

\usepackage{wrapfig}
\definecolor{lightblue}{rgb}{0.22,0.45,0.70}%
\usepackage{tabularx}
\usepackage{multirow}
\usepackage{adjustbox}

\usepackage{xcolor}
\usepackage{colortbl}
\definecolor{colgray}{RGB}{240,240,240}
\usepackage{pifont}

\definecolor{rliableolive}{HTML}{BBCC33}
\definecolor{rliableblue}{HTML}{77AADD}
\definecolor{rliablered}{HTML}{EE8866}

\usepackage{crossreftools}
\usepackage[suppress]{color-edits}

\usepackage{dsfont}
\usepackage{nicefrac}
\newtcolorbox{analysisbox}[1][]{
    enhanced jigsaw,
    colback=white,
    colframe=blue!75!black,
    fonttitle=\bfseries,
    boxsep=5pt,
    left=5pt,
    right=5pt,
    top=5pt,
    bottom=5pt,
    title=#1,
}
\definecolor{editInitialResponse}{RGB}{255, 235, 156} %
\definecolor{editBacktrack}{RGB}{0, 0, 139} %
\definecolor{editRevisedResponse}{RGB}{255, 182, 193} %

\usepackage{pifont}
\usepackage{soul}
\definecolor{highlightmistake}{RGB}{255, 179, 179} 
\definecolor{highlightcorrect}{RGB}{179, 255, 179}

\usepackage[capitalize,noabbrev]{cleveref}

\theoremstyle{plain}
\newtheorem{theorem}{Theorem}[section]

\theoremstyle{definition}

\newtheorem{definition}[theorem]{Definition}
\theoremstyle{remark}

\newtcolorbox{solutionbox}{
  colframe=black,
  colback=gray!10,
  boxrule=1pt,
  arc=0pt,
  title=,
  fonttitle=\bfseries
}
\newenvironment{sol}
  {\begin{solutionbox}}
  {\end{solutionbox}}
\newcommand{\BeginSol}{\begin{sol}}
\newcommand{\EndSol}{\end{sol}}

\usepackage[textsize=tiny]{todonotes}

\usepackage{enumitem}

\usepackage{amsmath,amsfonts,bm}

\def\eqref#1{Eq.~\ref{#1}}

\def\1{\bm{1}}

\DeclareMathAlphabet{\mathsfit}{\encodingdefault}{\sfdefault}{m}{sl}
\SetMathAlphabet{\mathsfit}{bold}{\encodingdefault}{\sfdefault}{bx}{n}

\definecolor{codegray}{gray}{0.9}
\definecolor{codepurple}{rgb}{0.58,0,0.82}
\definecolor{codeblue}{rgb}{0.25,0.5,0.5}

\lstdefinelanguage{YAML}{
  morekeywords={selector, sequence, condition, task, no},   %
  keywordstyle=\color{codeblue}\bfseries,                %
  ndkeywords={},                                        %
  sensitive=false,                                       %
  comment=[l]{\#},                                      %
  morecomment=[s]{/*}{*/},                               %
  commentstyle=\color{dkgreen}\ttfamily,               %
  string=[b]",                                          %
  stringstyle=\color{codepurple}\ttfamily,              %
  morestring=[b]',                                         %
  morestring=[b]`,                                         %
  identifierstyle=\ttfamily,                             %
  backgroundcolor=\color{codegray},                      %
  basicstyle=\ttfamily\footnotesize,
  breaklines=true,                                      %
  captionpos=b,                                         %
  frame=single,                                        %
  numbers=left,                                        %
  numberstyle=\tiny\color{gray},                        %
  numbersep=5pt,
  tabsize=2,                                           %
  showspaces=false,                                      %
  showstringspaces=false,                                %
  showtabs=false,                                        %
  xleftmargin=1em,
}
\renewcommand{\titlefont}{
  \color{BerkeleyBlue}
  \normalfont\bfseries
  \fontsize{15}{16}\selectfont
}
\title{Delegation Intelligence in Deep Search: A Controllable Framework for Disentangled Capability Diagnosis}

\author[1,2*]{Xinhao Yao}
\author[1*]{Yuanzhuo Liu}
\author[3]{Changhao Wang}
\author[4]{Yunfei Yu}
\author[2]{Haoran Tan}
\author[2]{Yuyao Zhang}
\author[2]{Ruifeng Ren}
\author[1]{Minlong Peng}
\author[2]{Yong Liu}

\affil[]{Tencent}
\affil[2]{Renmin University of China}
\affil[3]{UCAS}
\affil[4]{Independent Researcher}

\affil[*]{Equal contribution.}

\correspondingauthor{liuyonggsai@ruc.edu.cn\\   Our code and data will be available after internal review.}

\begin{document}
\maketitle
\textbf{Abstract:} 
Deep search is becoming a core capability of modern agent systems, yet it is typically evaluated solely based on end-to-end answer accuracy. This coupled evaluation paradigm entangles retrieval quality, long-context comprehension, evidence verification, and tool-use decisions, making it difficult to determine whether a model truly knows \emph{when} and \emph{how} to delegate information seeking to search. To this end: \ding{182} We formalize this meta-capability as \textit{\textbf{Delegation Intelligence}} in deep search and decompose it into complementary dimensions—\textit{Search Decision-Making} (recognizing information insufficiency and deciding whether, when, and how to search) and \textit{Information Synthesis \& Verification} (aggregating evidence from multiple sources, judging source reliability, and synthesizing information under noisy, potentially adversarial conditions). \ding{183} To enable disentangled and reproducible measurement, we develop a \textit{\textbf{controllable synthesis pipeline}} built on document-grounded reverse engineering. This yields a general recipe for constructing controlled deep-search evaluations rather than a single fixed dataset. \ding{184} As a concrete instantiation, we construct \textbf{\textit{DelegSearchBench}}, together with a \textit{disentangled evaluation protocol} that isolates each capability dimension by varying document composition and tool access. \ding{185} Across representative models, we demonstrate that deep-search competence cannot be adequately characterized by final-answer accuracy alone. Models exhibit different mode-dependent weaknesses, substantial sensitivity to the position of supporting evidence even when all relevant evidence is available, and unstable reasoning trajectories across repeated attempts. Under partial-context settings, many models either answer prematurely despite insufficient evidence or over-rely on search while failing to retrieve and integrate the information required to reach the correct answer. These results validate the controllable synthesis pipeline as a principled means of \textit{exposing capability boundaries masked by standard end-to-end evaluations}, showing that reliable deep-search agents require more than strong reasoning—they must also know how to reach correct answers through effective delegation.

\section{Introduction}
Over the past few years, language models have evolved into remarkably capable problem solvers, achieving strong performance on increasingly demanding mathematical, coding, and professional-knowledge benchmarks \citep{rein2024gpqa,ren2025deepseekproverv2,kimiteam2026kimik2openagentic,glm5team2026glm5vibecodingagentic}. Yet such achievements primarily characterize models as powerful \emph{test-takers}: they reward solving well-specified problems when the relevant information is already present in the prompt or encoded in model parameters \citep{dou2026clbenchbenchmarkcontextlearning}. Real-world information needs are rarely so self-contained. Evidence may be absent, scattered across sources, outdated, or entangled with plausible misinformation. A useful agent must therefore do more than reason over what it already has---it must recognize what it does not know and acquire the evidence needed to resolve that uncertainty. Deep search represents this transition from test-taking to active information seeking, extending static multi-hop reasoning \citep{hotpotqa,2WikiMultiHopQA,MuSiQue} to iterative document navigation and open-web investigation \citep{wu2025webwalker,chen2025xbench,wei2025browsecomp}. The central question is thus no longer only \emph{whether a model can produce the correct answer}, but \emph{whether it can reach that answer through sound information-seeking decisions}.

    Existing evaluations largely leave this question unanswered. Most deep-search benchmarks score the final response of an integrated system in which the model, retrieval backend, and tool interface are tightly coupled \citep{mialon2024gaia,wei2025browsecomp,browsecomp-plus}. A wrong answer may result from retrieval failure, poor long-context comprehension, misplaced trust in a source, or an incorrect decision not to search. Conversely, a correct answer may come from parametric memory, a lucky retrieval, or repeated sampling rather than effective search behavior. Collapsing these distinct trajectories into a single success rate makes failures difficult to diagnose and model capabilities difficult to compare. More fundamentally, it misses the defining challenge of deep search: \textit{deciding whether the available evidence is sufficient, determining when additional search is warranted, expressing the missing information need, and verifying the evidence returned}.

    We formalize this meta-capability as \textit{\textbf{Delegation Intelligence}} \citep{tomašev2026intelligentaidelegation,song2024exploring, kobis2025delegation,ning2026searchswarmdelegationintelligenceagentic} in deep search (Section \ref{sec:delegation}): the ability to make calibrated decisions about when and how information seeking should be delegated to search. Rather than treating retrieval as an unconditional preprocessing step, Delegation Intelligence views search as a \textbf{cognitive resource whose use depends on the current evidence state}. We decompose it into two complementary dimensions. Search Decision-Making measures whether a model recognizes evidence insufficiency and, when necessary, initiates and formulates an effective search. Information Synthesis \& Verification measures whether it can integrate evidence across documents, assess source authority and timeliness, detect inconsistencies, and reject plausible distractors. Separating information acquisition from post-retrieval reasoning enables attributable capability diagnosis.

    Disentangling these capabilities requires precise control over the evidential environment: an evaluator must know not only the correct answer, but also which evidence is necessary, which documents are merely irrelevant, and why seemingly plausible alternatives should be rejected. To this end, we develop a \textbf{controllable synthesis pipeline} (Section \ref{sec:pipieline}) based on document-grounded reverse engineering. Starting from high-quality real-world documents as complete evidential hindsight, the pipeline works backward to synthesize a natural query, its answer, localized supporting evidence, and the required reasoning structure. It preserves topically related retrieved documents as natural noise, constructs evidence-anchored distractors by perturbing query-constrained dimensions such as source authority, publication time, entity, and version, and applies multi-step rejection filtering to ensure answerability, non-memorability, evidence-label correctness, and adversarial validity. By fixing the answer and its evidential structure before task construction, the pipeline turns document composition into an explicit experimental variable rather than an uncontrolled consequence of retrieval. It therefore provides a reusable recipe for constructing diagnostic deep-search evaluations, rather than merely producing a single benchmark. As a concrete instantiation, we introduce \textit{\textbf{DelegSearchBench}},  and a disentangled protocol that varies document composition and tool availability.

     Our experiments (Section \ref{sec:eval}) reveal several capability boundaries hidden by endpoint accuracy. (1) Synthesis and verification are strongly mode-dependent: source authority, timeliness, internal consistency, multi-hop reasoning, and evidence sufficiency induce distinct model rankings and failure profiles. (2) When all required evidence is already available, most models remain sensitive to its position, with valid information buried among distractors often producing the largest degradation---a retrieval-independent manifestation of the lost-in-the-middle effect \citep{lost}. (3) Consistent gaps between Pass@1 and Pass@3 reveal unstable evidence localization and reasoning trajectories, echoing broader concerns about agent reliability across repeated attempts \citep{yao2025taubench}. Most importantly, partial-context evaluation uncovers two opposing delegation failures: some models answer prematurely despite missing evidence, while others search aggressively but fail to retrieve or integrate what is needed. 
     
     These results lead to our \textbf{central insight}: while search may be necessary, the mere invocation of search does not constitute intelligence. Reliable deep search requires well-calibrated coordination among uncertainty awareness, evidence acquisition, and faithful synthesis. These findings highlight the importance of capability-decoupled evaluation and suggest that future agent benchmarks should assess not only task-level performance, but also the robustness, reliability, and stability of the underlying reasoning processes.

\section{Related Work}
\textbf{Deep Information Seeking Tasks.} 
\ding{172} For deep search scenarios that demand reasoning across dispersed evidence, multi-hop reasoning datasets form the foundational line of evaluation. This category covers representative knowledge-intensive multi-hop QA benchmarks: 2WikiMultihopQA \citep{2WikiMultiHopQA}, MuSiQue \citep{MuSiQue}, Bamboogle \citep{Bamboogle}, and HotpotQA \citep{hotpotqa}. Constructed from structured knowledge bases or curated text corpora, these benchmarks require strict multi-step reasoning. However, they are predominantly derived from Wikipedia-based sources, which are heavily represented in conventional LLM pre-training data. \ding{173} A growing body of work extends evaluation to navigable document environments and general agent scenarios, aiming for more realistic deep information-seeking and task completion capabilities. For pure information-seeking objectives, document-navigation benchmarks represented by WebWalkerQA \citep{wu2025webwalker}, XBench \citep{chen2025xbench}, and Frames \citep{Frames} introduce iterative retrieval dynamics into multi-hop evaluation. Moving beyond information acquisition alone, broader general web agent benchmarks (GAIA \citep{mialon2024gaia}, HLE \citep{hle}) expand the scope to composite task completion. Compared with static multi-hop QA, these benchmarks more closely approximate real-world search behavior and agent workflows by incorporating iterative retrieval, path selection, and multi-tool coordination. However, they uniformly evaluate end-to-end system performance into a single task success rate, failing to disentangle retrieval backend quality and general task competence from a model’s intrinsic search-delegation abilities. \ding{174} To robustly evaluate deep search systems that require complex reasoning and strategic search planning, benchmarks centered on browsing-based information seeking are essential. BrowseComp \citep{wei2025browsecomp} and its extensions \citep{zhou2025browsecompzh,chen2025medbrowsecomp} are representative benchmarks explicitly designed for this purpose, featuring complex fact-finding queries with verifiable answers in a live web environment. \ding{175} Existing deep browsing benchmarks, however, primarily focus on question-answering performance of integrated systems without standardized fixed corpora, complicating fair comparative assessments of retrieval methodologies and agent decision-making. BrowseComp-Plus \citep{browsecomp-plus} addresses this issue by providing a human-verified static document corpus, enabling more controlled and reproducible evaluation. Even so, no existing work systematically disentangles the core metacognitive dimensions of search delegation — including when to search, which sources to trust, and how to verify evidence — leaving a critical gap in fine-grained capability diagnosis.

\noindent\textbf{Evaluation Focus.} In the early era of question answering and text generation, task contexts were relatively static and well-defined, making output validation the primary focus of evaluation. Evaluation methods evolved from exact-match metrics and human judgments to the widely adopted LLM-as-a-Judge paradigm \citep{li-etal-2025-generation,gu2025surveyllmasajudge}. However, \citet{dou2026clbenchbenchmarkcontextlearning,dou2026clbenchlifelanguagemodels} highlight a mismatch between current model training paradigms and real-world applications: the models we optimize are adept at reasoning over what they “already know,” whereas what is truly required is the ability to solve tasks grounded in messy, dynamically evolving context. Besides, output-level verification can only assess superficial plausibility of results, without tracing the compliance or correctness of the underlying execution process, and thus fails to detect reward hacking, shortcut behaviors, or latent policy violations \citep{metr-2025-recent-reward-hacking,macdiarmid2025naturalemergentmisalignmentreward}. To address limited process visibility, Claw-Eval \citep{ye2026claweval} shifts evaluation from final outputs to full execution trajectories. To capture inherent stochasticity in agent behavior, $\tau$-bench \citep{yao2025taubench} proposes a dual-metric framework—Pass@k and Pass all k—characterizing, respectively, the upper bound of capability and the lower bound of consistency. Meanwhile, the timeliness bias of static benchmarks has also drawn attention \citep{li2026clawevallive,zhu2026gisa}. Most agent benchmarks are released once with fixed task sets, which gradually drift away from real-world automation needs as tool ecosystems and business requirements evolve \citep{jain2025livecodebench,deng2026evoclaw}.

\section{Delegation Intelligence in Deep Search}\label{sec:delegation}
In this work, we argue that the essence of search agent intelligence lies not in a model's ability to answer queries correctly using only its parametric knowledge, but in its ability to make sound decisions that lead to efficient and effective use of search tools. To formalize this meta-competency for systematic evaluation, we define:

\begin{definition}
Delegation Intelligence in deep search is a model's metacognitive capacity to use search tools in open environments — autonomously gauging information needs, choosing cognitive routes, and vetting source reliability to achieve efficient acquisition and reasoning. Its core lies not in whether the model can produce the correct answer on its own, but in \textbf{whether the model can make the right decisions about tool use: when to search, how to search, which sources to trust, and which reasoning path to follow}.
\end{definition}

To operationalize this concept for empirical evaluation, we decompose Delegation Intelligence in deep search into two mutually complementary capability dimensions. Each dimension encompasses a set of fine-grained sub-capabilities:

\begin{itemize}

    \item \textbf{Search Decision-Making} governs both the high-level selection of an overall problem-solving strategy and the tactical decisions for search execution. It first assesses whether the available contextual evidence is sufficient to answer a query. When the context is adequate, the model reasons directly from it; otherwise, it either initiates external web search to acquire or verify the required information or relies on parametric knowledge when retrieval is unnecessary. If web search is selected, it further decomposes complex queries into independently retrievable subproblems and reformulates ambiguous or underspecified queries to improve retrieval precision.
    \item \textbf{Information Synthesis \& Verification} concerns post-retrieval information processing. It encompasses three key capabilities: (i) evaluating source reliability using authority signals, publication timeliness, and cross-source consensus; (ii) integrating evidence from multiple documents; and (iii) identifying internal inconsistencies within individual documents.

\end{itemize}

While this two-dimensional taxonomy defines the scope of Delegation Intelligence, most existing deep search benchmarks evaluate retrieval, reading comprehension, and reasoning as a tightly coupled end-to-end process. Such conflation prevents capability-specific attribution of model performance and renders the evaluation highly susceptible to retrieval-induced noise, thereby obscuring the model's intrinsic reasoning and tool-use competence.

To enable disentangled and fine-grained evaluation, we provide the complete message context—namely, the query together with a controlled document set containing supporting evidence, misleading distractors, and irrelevant background documents. This \textit{\textbf{controlled setting}} \citep{cao2025ctrlsynth,wu-etal-2025-trident,huang2026building,cheng-etal-2026-beyond} isolates the model's intrinsic reasoning and tool-use capabilities by eliminating confounding variance from external retrieval, shifting the evaluation focus from end-to-end system performance to intrinsic decision-making and information processing. Details of the construction are presented in the next section.
\section{Controllable Synthesis Pipeline}\label{sec:pipieline}
\begin{figure*}[htb]
\centering
\includegraphics[width=\textwidth]{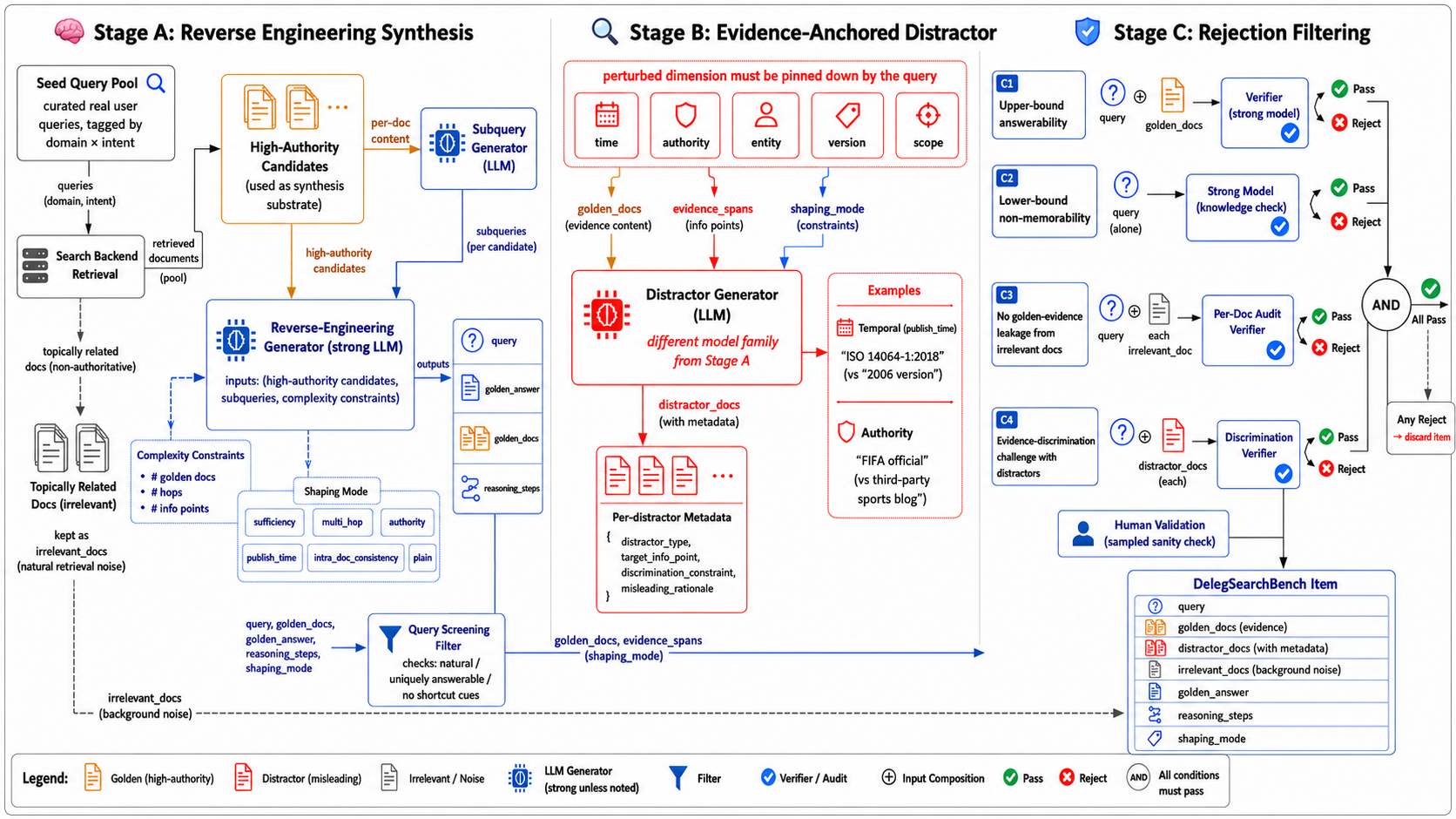}
\caption{\textbf{Overview of the controllable synthesis pipeline: }reverse-engineered query synthesis, evidence-aligned distractor generation, and multi-criteria rejection filtering. Items passing all automated filters enter the candidate benchmark pool,
whose residual quality is subsequently assessed through a stratified
human audit. Irrelevant retrieval documents from Stage A are preserved as natural background noise for search evaluation. See the main text for details.}
\label{fig:pipeline}
\end{figure*}
To realize this controlled evaluation setting, we develop a multi-stage synthesis pipeline (Figure \ref{fig:pipeline}) based on document-grounded reverse engineering. The key idea is to leverage high-quality real-world documents—such as technical reports, official announcements, and policy documents—as complete hindsight for benchmark construction. Unlike conventional forward-generation paradigms, which first formulate a query and subsequently retrieve or curate supporting evidence, our pipeline starts from authentic documents and works backward to synthesize the corresponding user query. During this process, the ground-truth answer, supporting evidence, reasoning structure, and targeted capability under evaluation are all explicitly specified in advance.

This reverse design is crucial for controllability. Query-first pipelines often rely on incomplete, ambiguous, or unavailable evidence, and their evaluation results are therefore confounded by retrieval backend performance. In contrast, grounding task generation in finalized high-quality documents provides full prior knowledge of both the target answer and its evidential structure before task formulation, enabling more controlled and reliable evaluation.

\subsection{Stage A: Reverse Engineering Synthesis}
Stage A starts from a curated seed-query pool (see Appendix \ref{app:pipe} for domain details.) rather than arbitrary prompts. These seeds are selected from high-quality real user queries and organized across broad domains and intent types, providing diverse entry points while keeping the upstream distribution aligned with realistic search behavior. 

Given a seed query, we first call a search backend to retrieve a pool of relevant documents. \textbf{High-authority candidate documents (e.g., official reports) are then used as the synthesis substrate, while the remaining topically related documents are retained as natural retrieval noise} (\texttt{irrelevant\_docs}). For each high-authority candidate, we generate subqueries that reveal what information needs the document can support. Finally, a strong generator model synthesizes across the candidate documents and subqueries to produce a natural user query, a concise golden answer, supporting evidence in \texttt{golden\_docs}, and reasoning steps. In this way, the final query and answer are reverse-engineered \citep{li2025reportbench,hu2025stepdeepresearch,lu2026search} from verified evidence rather than freely hallucinated.

We additionally apply lightweight constraints on the number of golden documents, reasoning hops, and independent information points to control sample complexity. The synthesized query is screened to ensure that it is natural, uniquely answerable, and free of shortcut cues such as copied answer tokens or source-specific identifiers. Concrete prompts, model configuration and representative examples are deferred to Appendix \ref{app:stageA}. Stage A also assigns a shaping mode, which specifies the capability axis to be stressed and provides structural guidance for Stage B distractor generation:

\(\bullet\) \texttt{sufficiency}: parallel aggregation across multiple golden documents; 

\(\bullet\) \texttt{multi\_hop}: serial reasoning where one hop feeds into the next; 

\(\bullet\) \texttt{authority}: source-reliability judgment using visible source cues such as site\_name, url, and writing style; 

\(\bullet\) \texttt{publish\_time}: temporal reliability judgment based on publication time; 

\(\bullet\) \texttt{intra\_doc\_consistency}: consistency judgment within a single document; 

\(\bullet\) \texttt{plain}: no extra capability shaping, used as a control condition.

This reverse-engineering design ensures: \ding{182} \textbf{answerability}, as each item is grounded in localized evidence and supported by contributing golden documents; \ding{183} \textbf{non-triviality}, as tasks are derived from realistic information-seeking structures rather than shallow templates; and \ding{184} \textbf{retrieval realism}, achieved by grounding synthesis in high-authority documents while retaining topically related retrieved documents as natural retrieval noise to preserve the ambiguity and diversity of real-world search scenarios.

\subsection{Stage B: Evidence-Anchored Distractor}
While the retrieved \texttt{irrelevant\_docs} preserve natural background noise, they are insufficient for controlled evaluation of distraction discrimination: their misleading effect is incidental and may not target the evidence required by the golden answer. Stage B therefore uses a generator from a different model family to synthesize adversarial \texttt{distractor\_docs} from the Stage-A supporting evidence. A valid distractor \textbf{must satisfy} a structural constraint: the dimension along which it differs from the golden evidence must be a dimension already pinned down by the query itself. Each distractor records its target information, distractor type, and discriminating constraint, making the source of distraction explicit and auditable. Refer to Appendix \ref{app:stageB} for detailed implementation settings.

For example, consider the query “What is the organizational carbon accounting boundary specified in ISO 14064-1:2018?”. A document describing the 2006 version of the same standard acts as a strong temporal distractor: the query explicitly pins the standard to its 2018 revision, so a rigorous model can identify and reject the outdated specification by cross-checking the version dimension. For an authority-focused case, take the query “According to FIFA’s official statistics, how many teams qualified for the 2026 FIFA World Cup?”. A third-party sports blog claiming 32 teams serves as a plausible but invalid distractor. Since the query explicitly anchors the data source to official FIFA statistics, the distractor deviates exactly on the source authority dimension that the query has already specified, rather than introducing unrelated factual errors.
In both cases, distractors remain topically aligned and pass shallow keyword matching, but fail strict evidence verification. This design produces targeted, interpretable challenge samples that map cleanly to the synthesis modes defined in Stage A.

\subsection{Stage C: Rejection Filtering}
Stage C serves as a rejection filter rather than a final evaluation protocol. Its purpose is to remove structurally flawed synthesized items before they enter the benchmark pool. Each filtering step runs under a \textbf{different controlled context}, so the context itself determines what property is being tested.

The filtering procedure consists of four views, where \(\oplus\) denotes input composition. \ding{172} \textbf{C1} uses \texttt{query} \(\oplus\) \texttt{golden\_docs} to test the upper-bound answerability of the item: if a verifier cannot recover the golden answer even when given the intended evidence, the query is ambiguous, the answer is wrong, or the evidence spans are insufficient, and the item should be rejected. \ding{173} \textbf{C2} uses \texttt{query} alone, with no documents, to test the lower-bound non-memorability of the item: if strong models can recover the  golden answer from the query alone, the item is considered susceptible to parametric knowledge leakage rather than requiring retrieval, and is therefore rejected. \ding{174} \textbf{C3} audits each \texttt{irrelevant\_doc} with query independently to ensure that documents labeled as irrelevant contain no supporting golden evidence. \ding{175} \textbf{C4} uses \texttt{query} \(\oplus\) \texttt{distractor\_docs}, to assess whether the distractors create a meaningful evidence-discrimination challenge. A useful distractor need not simply make models answer incorrectly; rather, it should force the model to rely on the correct evidence, inspect the query constraints, and perform retrieval-grounded reasoning instead of accepting superficially relevant documents. Implementation details are provided in Appendix \ref{app:stageC}.

Together, these stages ensure that the benchmark pool contains only items whose \textbf{answerability, non-memorability, evidence labeling, and adversarial strength} have passed explicit rejection criteria.

\noindent $\diamond$ \textbf{Human Validation.} Although the construction pipeline is largely automated, we conduct a lightweight, stratified post-construction audit as a final quality assurance measure. Importantly, human validation is primarily diagnostic rather than an exhaustive item-by-item acceptance or re-annotation stage: it provides an independent sanity check by estimating residual defect rates in the Stage-C-accepted candidate pool. The small number of items flagged as defective by the human majority are discarded from the final released pool without manual correction or re-annotation. We sample accepted items across domains, modes, and complexity levels to identify potential failure patterns and assess human confirmation of the primary quality labels.  Detailed audit statistics are reported in Appendix~\ref{app:human}.
\section{Disentangled Evaluation Protocol}\label{sec:eval}

\subsection{Overview and Settings}
\noindent $\diamond$ \textbf{Benchmark Overview.} Building on the controllable synthesis pipeline, we construct \textbf{{DelegSearchBench}}, a benchmark dedicated to the disentangled evaluation of Delegation Intelligence in deep search. The final pool comprises 429 items produced by the pipeline described above, spanning six capability modes. Detailed data distribution information is provided in the appendix. Building on this benchmark, we introduce a disentangled evaluation paradigm that independently measures the two core dimensions of Delegation Intelligence—\textbf{Search Decision-Making} and \textbf{Information Synthesis \& Verification}—without confounding from external retrieval systems. By systematically varying the composition of the input document set and the availability of tool access, we isolate each capability dimension and obtain interpretable, attributable performance measurements.

 \begin{table}[htb]
    \centering
    \fontsize{6}{6}\selectfont 
    \setlength{\tabcolsep}{2pt} 
    \begin{tabular}{llrrrrrr}
    \toprule
    \textbf{Mode} & \textbf{Model} & \multicolumn{2}{c}{\textbf{golden\_first}} & \multicolumn{2}{c}{\textbf{golden\_mid}} & \multicolumn{2}{c}{\textbf{golden\_last}} \\
    \cmidrule(lr){3-4}\cmidrule(lr){5-6}\cmidrule(lr){7-8}
     & & P@1 & P@3 & P@1 & P@3 & P@1 & P@3 \\
    \midrule
    \texttt{authority} & \texttt{doubao-seed-2.0-pro} & 0.624 & 0.742 & 0.452 & 0.516 & 0.548 & 0.677 \\
     & \texttt{qwen3.7-plus} & 0.753 & 0.806 & 0.731 & 0.871 & 0.806 & \textbf{0.968} \\
     & \texttt{hunyuan-3-preview} & 0.731 & 0.871 & 0.602 & 0.742 & 0.624 & 0.742 \\
     & \texttt{glm-5.1} & 0.828 & 0.903 & 0.710 & 0.806 & 0.720 & 0.806 \\
     & \texttt{minimax-m2.7} & 0.871 & 0.903 & 0.688 & 0.806 & 0.699 & 0.839 \\
     & \texttt{deepseek-v4-pro} & 0.796 & \textbf{0.968} & 0.634 & 0.774 & 0.688 & 0.903 \\
     & \texttt{kimi-k2.6} & 0.871 & 0.935 & 0.731 & 0.806 & 0.763 & 0.839 \\
     & \texttt{gemini-3.1-pro-preview} & \textbf{0.925} & 0.968 & \textbf{0.935} & \textbf{1.000} & \textbf{0.892} & 0.935 \\
     & \texttt{gpt-5-mini} & 0.828 & 0.839 & 0.742 & 0.839 & 0.699 & 0.774 \\
    \midrule
    \texttt{intra\_doc.} & \texttt{doubao-seed-2.0-pro} & 0.521 & 0.649 & 0.504 & 0.628 & 0.443 & 0.617 \\
     & \texttt{qwen3.7-plus} & 0.667 & 0.840 & 0.603 & 0.745 & 0.642 & 0.819 \\
     & \texttt{hunyuan-3-preview} & 0.585 & 0.713 & 0.539 & 0.649 & 0.543 & 0.702 \\
     & \texttt{glm-5.1} & 0.684 & 0.787 & 0.635 & 0.755 & \textbf{0.702} & \textbf{0.819} \\
     & \texttt{minimax-m2.7} & 0.684 & 0.840 & 0.578 & 0.723 & 0.592 & 0.745 \\
     & \texttt{deepseek-v4-pro} & 0.684 & 0.862 & 0.543 & 0.745 & 0.557 & 0.681 \\
     & \texttt{kimi-k2.6} & \textbf{0.745} & \textbf{0.883} & 0.635 & \textbf{0.809} & 0.649 & 0.798 \\
      & \texttt{gemini-3.1-pro-preview} & 0.713 & 0.809 & 0.635 & 0.755 & 0.621 & 0.691 \\
      & \texttt{gpt-5-mini} & 0.691 & 0.819 & 0.571 & 0.723 & 0.589 & 0.734 \\
    \midrule
    \texttt{multi\_hop} & \texttt{doubao-seed-2.0-pro} & 0.603 & 0.709 & 0.511 & 0.646 & 0.527 & 0.658 \\
     & \texttt{qwen3.7-plus} & 0.709 & 0.848 & 0.624 & 0.734 & 0.654 & 0.759 \\
     & \texttt{hunyuan-3-preview} & 0.700 & 0.835 & 0.608 & 0.722 & 0.650 & 0.785 \\
     & \texttt{glm-5.1} & 0.751 & 0.899 & 0.700 & 0.785 & 0.726 & 0.835 \\
     & \texttt{minimax-m2.7} & 0.776 & 0.873 & 0.654 & 0.785 & 0.671 & 0.810 \\
     & \texttt{deepseek-v4-pro} & 0.789 & \textbf{0.924} & 0.641 & 0.785 & 0.671 & 0.861 \\
     & \texttt{kimi-k2.6} & \textbf{0.843} & 0.911 & 0.714 & 0.846 & 0.731 & \textbf{0.885} \\
     & \texttt{gemini-3.1-pro-preview} & 0.759 & 0.873 & \textbf{0.759} & \textbf{0.873} & \textbf{0.785} & 0.873 \\
     & \texttt{gpt-5-mini} & 0.797 & 0.899 & 0.620 & 0.734 & 0.688 & 0.797 \\
    \midrule
    \texttt{plain} & \texttt{doubao-seed-2.0-pro} & 0.429 & 0.548 & 0.349 & 0.476 & 0.405 & 0.595 \\
     & \texttt{qwen3.7-plus} & 0.563 & 0.667 & 0.476 & 0.643 & 0.532 & 0.690 \\
     & \texttt{hunyuan-3-preview} & 0.500 & 0.619 & 0.500 & 0.571 & 0.532 & 0.595 \\
     & \texttt{glm-5.1} & 0.627 & 0.667 & \textbf{0.603} & 0.690 & \textbf{0.603} & 0.714 \\
     & \texttt{minimax-m2.7} & 0.516 & 0.667 & 0.500 & 0.667 & 0.476 & 0.643 \\
     & \texttt{deepseek-v4-pro} & \textbf{0.675} & \textbf{0.833} & 0.508 & 0.714 & 0.484 & 0.643 \\
     & \texttt{kimi-k2.6} & 0.595 & 0.714 & 0.540 & 0.738 & 0.524 & 0.643 \\
     & \texttt{gemini-3.1-pro-preview} & 0.627 & 0.762 & 0.603 & \textbf{0.786} & \textbf{0.619} & \textbf{0.762} \\
    & \texttt{gpt-5-mini} & 0.595  &  0.667  &  0.468  &  0.595  &  0.476  & 0.571 \\
    \midrule
    \texttt{publish\_time} & \texttt{doubao-seed-2.0-pro} & 0.597 & 0.698 & 0.481 & 0.605 & 0.473 & 0.523 \\
     & \texttt{qwen3.7-plus} & 0.643 & 0.756 & 0.570 & 0.709 & 0.632 & 0.779 \\
     & \texttt{hunyuan-3-preview} & 0.651 & 0.767 & 0.539 & 0.640 & 0.570 & 0.686 \\
     & \texttt{glm-5.1} & 0.705 & 0.791 & 0.663 & 0.767 & 0.709 & 0.802 \\
     & \texttt{minimax-m2.7} & 0.659 & 0.791 & 0.504 & 0.663 & 0.562 & 0.686 \\
     & \texttt{deepseek-v4-pro} & 0.643 & 0.791 & 0.504 & 0.686 & 0.523 & 0.651 \\
     & \texttt{kimi-k2.6} & 0.690 & 0.826 & 0.585 & 0.744 & 0.620 & 0.733 \\
          & \texttt{gemini-3.1-pro-preview} & \textbf{0.822} & \textbf{0.895} & \textbf{0.736} & \textbf{0.837} & \textbf{0.748} & \textbf{0.884} \\
      & \texttt{gpt-5-mini} & 0.616 & 0.721 & 0.430 & 0.535 & 0.504 & 0.593 \\
    \midrule
    \texttt{sufficiency} & \texttt{doubao-seed-2.0-pro} & 0.436 & 0.557 & 0.344 & 0.474 & 0.330 & 0.433 \\
     & \texttt{qwen3.7-plus} & 0.625 & 0.742 & 0.540 & 0.680 & 0.557 & 0.701 \\
     & \texttt{hunyuan-3-preview} & 0.495 & 0.639 & 0.402 & 0.546 & 0.409 & 0.515 \\
     & \texttt{glm-5.1} & 0.649 & 0.773 & 0.574 & 0.691 & 0.581 & 0.691 \\
     & \texttt{minimax-m2.7} & 0.622 & 0.742 & 0.512 & 0.649 & 0.491 & 0.649 \\
     & \texttt{deepseek-v4-pro} & 0.605 & 0.753 & 0.471 & 0.649 & 0.447 & 0.649 \\
     & \texttt{kimi-k2.6} & 0.701 & \textbf{0.876} & 0.619 & 0.794 & 0.567 & 0.711 \\
       & \texttt{gemini-3.1-pro-preview} & \textbf{0.742} & 0.866 & \textbf{0.708} & \textbf{0.814} & \textbf{0.718} & \textbf{0.784} \\
    & \texttt{gpt-5-mini} & 0.625  & 0.742  & 0.522  & 0.649  &  0.467  &  0.577 \\
    \bottomrule
    \end{tabular}
    \caption{Setting I results by shaping mode and model. For each evidence-position variant, we report Pass@1 and Pass@3. Best results per column are marked in \textbf{bold}.}
    \label{tab:setting1-results}
    \end{table}     

\textbf{Evaluation Settings.} Each item's document pool (golden $\cup$ distractor $\cup$ irrelevant) can be presented under two complementary settings that isolate the two complementary capability dimensions introduced in Section~\ref{sec:delegation}. Model outputs are judged by a held-out judge model deepseek-v4-flash that is never included in the target-model panel, avoiding direct self-evaluation. We synthesize data using recent GPT and Claude variants (gpt-5.2, claude-opus-4.6) and exclude evaluations of these models to avoid self-assessment bias.

\ding{172} \textbf{Setting I} (Full-Context, search disabled) presents the model with the \textbf{complete} document pool ($\texttt{golden\_docs} \cup \texttt{distractor\_docs} \cup \texttt{irrelevant\_docs}$) directly in a single turn, with no search tool exposed. Since all required evidence is already provided and no search decision is involved, this setting isolates \textbf{Information Synthesis \& Verification} by evaluating the model's ability to synthesize cross-document evidence, assess source authority and timeliness, detect inconsistencies, and disregard distractor and irrelevant documents. We further construct three document-order variants with identical document sets but different presentation orders to evaluate robustness to evidence position.

\ding{173} \textbf{Setting II} (Partial-Context, search enabled) provides only a deliberately incomplete subset of \texttt{golden\_docs}, omitting one or more evidence pieces required to fully answer the query, while retaining the complete \texttt{distractor\_docs} and \texttt{irrelevant\_docs} sets and granting access to a search tool. The model must first determine whether the available evidence is sufficient to answer the question directly, choosing between directly answer and further search. This setup is explicitly designed to isolate \textbf{Search Decision-Making}. Because the withheld gold evidence is genuinely absent from the initial context, rather than embedded within the provided documents, the model cannot compensate for missing information through more exhaustive reading of the available context. Consequently, performance serves as a direct measure of the key capabilities: (1) recognizing genuine information insufficiency, (2) decomposing or reformulating queries to effectively target the missing evidence, and (3) exercising calibrated judgment in determining whether external retrieval is necessary.

\subsection{Evaluation Results}

\noindent $\diamond$ \textbf{Setting I:} We first evaluate models under Setting I, where the full document pool is provided and search is disabled. \textit{This setting isolates models' ability to synthesize evidence and filter distractors without retrieval uncertainty}. Table~\ref{tab:setting1-results} reports the results for each shaping mode and model. We report both Pass@1 and Pass@3 under three evidence-position variants\footnote{Specifically, \texttt{golden\_first} places the golden documents at the front of the input sequence, preceding all distractor and irrelevant documents; \texttt{golden\_mid} inserts the golden document amid a mixture of noisy distractor texts; and \texttt{golden\_last} puts valid golden evidence after all distracting passages.}: \texttt{golden\_first}, \texttt{golden\_mid}, and \texttt{golden\_last}. This dual set of metrics enables joint analysis of overall model performance and positional sensitivity toward supporting evidence.

Several common patterns emerge from Table \ref{tab:setting1-results}: \ding{182} The \textbf{full-text reasoning performance of LLMs is highly mode-dependent}, a consistent pattern across all evaluated models. Each mode paradigm introduces a distinct form of document interference, requiring specialized reasoning capabilities such as source attribution, temporal reasoning and multi-source information integration. Variations in interference complexity and capability requirements produce clear performance stratification, indicating that \textbf{full-text information extraction and verification is not a homogeneous capability but a composite of multiple specialized reasoning competencies}. Consequently, no existing model consistently achieves strong performance across all mode paradigms. \ding{183} \textbf{Most models exhibit non-negligible sensitivity to the position of supporting evidence}, a pattern consistent with the well-documented lost-in-the-middle \citep{lost} phenomenon. Across all task modes, on average and for most models, performance consistently peaks when golden documents are placed at the front of the input context (\texttt{golden\_first}), and degrades substantially when evidence is embedded in the middle (\texttt{golden\_mid}) or shifted to the end (\texttt{golden\_last}). In most settings, the sharpest performance drop occurs in the \texttt{golden\_mid} condition, reflecting that models have the weakest ability to locate and utilize valid information buried among distractor passages. \textbf{Critically, this positional bias persists even in our controlled full-context setting with retrieval uncertainty eliminated, confirming that it is an inherent limitation of the model’s context processing mechanism rather than a byproduct of retrieval errors}. \ding{184} Pass@3 consistently exceeds Pass@1 across task modes and evidence positions, indicating substantial run-to-run variability in model performance. While repeated sampling increases the likelihood of obtaining a correct response, the observed gap suggests that many models do not reliably follow stable evidence-localization and reasoning trajectories across independent generations. \textbf{For a genuinely competent agent, however, decision paths should be stable and convergent. A model prone to missing key evidence or deviating from correct reasoning under stochastic sampling perturbations cannot be said to have truly mastered the task}. This instability is further amplified when golden evidence is not placed upfront, meaning multiple sampling only probabilistically alleviates the lost-in-the-middle effect, without resolving the inherent fragility of context utilization.

\begin{table}[htb]
\centering
\fontsize{7}{8}\selectfont
\setlength{\tabcolsep}{2pt}
\begin{tabular}{lrrrrrr}
\toprule
\textbf{Model} & Search & Direct & \cellcolor{colgray}\textbf{P@1}  & \cellcolor{colgray}\textbf{P@3}  
& \multicolumn{1}{c}{\shortstack{Hit@First\\Search}}
& \multicolumn{1}{c}{\shortstack{Correct@\\Search}} \\
\midrule
\multicolumn{7}{l}{\textbf{multi\_hop}} \\
\texttt{doubao-seed-2.0-pro} & 0.308 & 0.692 & 0.154 & 0.231 & 0.500 & 0.500 \\
\texttt{qwen3.7-plus} & 0.462 & 0.538 & 0.282 & 0.462 & 0.389 & 0.611 \\
\texttt{hunyuan-3-preview} & 0.872 & 0.128 & 0.436 & 0.538 & 0.206 & 0.500 \\
\texttt{glm-5.1} & 0.795 & 0.205 & 0.564 & 0.769 & 0.452 & 0.710 \\
\texttt{minimax-m2.7} & 0.590 & 0.410 & 0.282 & 0.538 & 0.478 & 0.478 \\
\texttt{deepseek-v4-pro} & 0.154 & \textbf{0.846} & 0.103 & 0.308 & 0.333 & 0.667 \\
\texttt{kimi-k2.6} & 0.385 & 0.615 & 0.359 & 0.385 & 0.933 & \textbf{0.933} \\
\texttt{gemini-3.1-pro-preview} & 0.718 & 0.282 & 0.538 & 0.769 & 0.750 & 0.750\\
\texttt{gpt-oss-120b} & 0.462 & 0.538  & 0.282 & 0.308 & 0.722 & 0.611 \\
\texttt{gpt-5-mini} & 0.615 &0.385 &0.308  & 0.462  &0.708  &0.500\\
\texttt{claude-sonnet-4.6}   &\textbf{0.941}  &0.059  & \textbf{0.824}   & \textbf{0.846}  & \textbf{0.938}   & 0.875 \\
\midrule
\multicolumn{7}{l}{\textbf{sufficiency}} \\
\texttt{doubao-seed-2.0-pro} & 0.050 & \textbf{0.950} & 0.000 & 0.000 & 0.333 & 0.000 \\
\texttt{qwen3.7-plus} & 0.433 & 0.567 & 0.150 & 0.250 & 0.500 & 0.346 \\
\texttt{hunyuan-3-preview} & 0.783 & 0.217 & 0.150 & 0.250 & 0.383 & 0.191 \\
\texttt{glm-5.1} & 0.467 & 0.533 & 0.250 & \textbf{0.500} & 0.464 & \textbf{0.536} \\
\texttt{minimax-m2.7} & 0.533 & 0.467 & 0.133 & 0.250 & 0.562 & 0.250 \\
\texttt{deepseek-v4-pro} & 0.317 & 0.683 & 0.200 & 0.350 & 0.684 & 0.368 \\
\texttt{kimi-k2.6} & 0.183 & 0.817 & 0.100 & 0.200 & 0.727 & 0.273 \\
\texttt{gemini-3.1-pro-preview} & 0.467 & 0.533& 0.283 &  0.350& \textbf{0.929} & 0.536\\
\texttt{gpt-oss-120b} & 0.283 & 0.717 &   0.067 & 0.100 & 0.471 & 0.235 \\
\texttt{gpt-5-mini} & 0.583 & 0.417& 0.233&0.450&0.657&0.343 \\
\texttt{claude-sonnet-4.6} & \textbf{0.895} & 0.105 & \textbf{0.474} & 0.500 & 0.647 & 0.529\\
\bottomrule
\end{tabular}
\caption{Setting II results for tasks requiring missing-evidence recovery. \textbf{Search}/\textbf{Direct}: fraction of samples where the model selects search / direct answering (action distribution);
\textbf{P@1}/\textbf{P@3}: overall Pass@1 and Pass@3 answer accuracy across all samples;
\textbf{Hit@FirstSearch}: hit rate of at least one withheld golden document in the model's first autonomous search, computed exclusively on samples with search actions;
\textbf{Correct@Search}: final answer accuracy conditional on the model having performed a search action.
The best result in each column is marked in \textbf{bold}.}
\label{tab:setting2-results}
\end{table}

\noindent $\diamond$ \textbf{Setting II:} We then evaluate models in a partial-context retrieval environment. Since this setting is designed to test whether a model can notice missing evidence and proactively search for it, we restrict the evaluation to the two modes: \texttt{multi\_hop} and \texttt{sufficiency}. For each item in this subset, the initial context contains only a deliberately selected subset of \texttt{golden\_docs}, omitting at least one gold document or evidence point required to fully answer the query. Meanwhile, all \texttt{distractor\_docs} and \texttt{irrelevant\_docs} are still included, and the model is given access to the search tool. Importantly, this is not an open-world web search setting: the search backend retrieves the top-$k$ documents from the current item's document pool according to the model's query. This choice keeps retrieval reproducible and prevents uncontrolled external-search variance, while still testing whether the model can decide to search and express the missing information need effectively. \textbf{This construction makes the missing evidence genuinely absent from the initial context rather than merely hidden among long passages, so the model cannot solve the task by simply reading more carefully. At the same time, retaining the full distractor and irrelevant document sets preserves the noisy evidence environment of realistic search}. Setting II therefore isolates search decision-making under controlled but non-oracle conditions: the model must recognize insufficiency, decide whether to call search, formulate a useful query, and integrate any retrieved evidence with the partial context already provided.

Table~\ref{tab:setting2-results} shows that this setting is substantially more challenging than full-context synthesis. \ding{182} Given the necessity of hidden evidence for task solving, models persist in resorting to direct response generation. \textbf{Premature answering continues to serve as a major failure mode, reflecting the models’ deficient ability to perceive inadequate evidence conditions}. \ding{183} \textbf{Search, however, is necessary but not sufficient}. Models further show distinct delegation profiles. \texttt{claude-sonnet-4.6} achieves a good final performance in both \texttt{multi\_hop} (Pass@1 0.824, Pass@3 0.846) and \texttt{sufficiency} (Pass@1 0.474, Pass@3 0.500), combining a relatively high search rate with effective post-retrieval synthesis. By contrast, \texttt{doubao-seed-2.0-pro} and \texttt{deepseek-v4-pro} frequently answer directly despite incomplete evidence, leading to high Direct rates and low final accuracy. \texttt{hunyuan-3-preview} exhibits the opposite behavior: it searches aggressively, but its lower Hit@Search and Correct@Search show that tool use alone is not enough. These results highlight that Delegation Intelligence requires calibrated search behavior: models must know when evidence is missing, how to retrieve the missing evidence, and how to integrate it with the noisy context already available.

\noindent $\diamond$ \textbf{Summary.} Taken together, the evaluation results demonstrate: (1) \textbf{End-to-end agent benchmarks are insufficient for capability diagnosis} because they conflate reasoning, retrieval, planning, and tool-use failures into a single success metric. We therefore advocate capability-decoupled evaluation under controlled settings. (2) Agent competence is inherently multi-dimensional. Different models exhibit distinct capability profiles, indicating that \textbf{similar overall accuracy can mask fundamentally different reasoning bottlenecks}. (3) \textbf{Long-horizon agent evaluation should emphasize not only effectiveness but also stability}. Large Pass@1–Pass@3 gaps suggest that current models often succeed through stochastic exploration rather than consistently reliable reasoning, making robustness across repeated executions an important evaluation objective.
\section{Conclusion}
    We introduced \textit{\textbf{Delegation Intelligence}} in deep search as a framework for evaluating deep-search agents beyond final-answer accuracy. We operationalized this framework through a controllable, document-grounded synthesis pipeline, and a protocol that varies evidence completeness, document order, and tool access. Our evaluation with \textit{\textbf{DelegSearchBench}} reveals capability boundaries obscured by end-to-end scores: models exhibit mode-specific weaknesses, positional sensitivity despite complete evidence, unstable reasoning across attempts, and two opposing delegation failures---premature answering and ineffective searching. Reliable deep search therefore requires more than invoking a tool; it requires calibrated coordination among recognizing uncertainty, acquiring appropriate evidence, and verifying and synthesizing that evidence faithfully.

\bibliography{main}

\newpage
\appendix
\onecolumn
\section{Pipeline Details}\label{app:pipe}

\textbf{Seed query domain.} As shown in Table~\ref{tab:domain-stats}, seed queries are drawn from several real-world domains inherited from the seed pool's intent taxonomy, led by finance (18.2\%), legal (14.0\%), shopping (12.4\%), entertainment (10.8\%), academic knowledge (10.3\%), and healthcare (9.9\%), with a long tail of news, politics, technology, travel, and lifestyle domains — reflecting a realistic, non-uniform distribution of real user search intents rather than an artificially balanced sample.

\begin{table}[htb]
\centering
\small
\begin{tabular}{lrr}
\toprule
Domain & \#Items & \% \\
\midrule
Finance \& Economics       & 79 & 18.2 \\
Legal                      & 61 & 14.0 \\
Shopping / Consumer Goods  & 54 & 12.4 \\
Entertainment              & 47 & 10.8 \\
Academic Knowledge         & 45 & 10.3 \\
Healthcare                 & 43 &  9.9 \\
News \& Current Events     & 25 &  5.7 \\
Politics \& Military       & 23 &  5.3 \\
AI \& Computer Science     & 20 &  4.6 \\
Other (long-tail domains)  & 32 &  8.7 \\
\bottomrule
\end{tabular}
\caption{Domain distribution of the current data pool. Long-tail minor domains are grouped into the ``Other'' category.}
\label{tab:domain-stats}
\end{table}

\subsection{Stage A Prompts and Examples}\label{app:stageA}

\textbf{Seed-query expansion prompt.}
When additional seeds are needed, we expand the curated seed pool in a domain-aware manner. The generator is given several real seed queries from the same broad domain and asked to produce new queries that preserve the same high-level properties:
\begin{quote}
\small
\textbf{Input}: domain label; several real seed-query examples from that domain.\\
\textbf{Instruction}: generate new self-contained factual search queries. The queries should be natural, domain-consistent, non-subjective, independent of previous context, free of relative-time expressions, and have a unique real-world answer. Cover different intent types when possible, such as factual lookup, causal analysis, comparison, and temporal information seeking.\\
\textbf{Output}: a list of candidate seed queries with lightweight domain and intent annotations.
\end{quote}

\textbf{Document-to-subquery prompt.}
After retrieving documents for a seed query, Stage A first analyzes each high-authority candidate document independently. The goal is not yet to produce the final benchmark query, but to expose what information needs the document can support:
\begin{quote}
\small
\textbf{Input}: one high-authority candidate document, including title, source, publication time, and content snippet.\\
\textbf{Instruction}: write several search subqueries that a real user might ask if they needed the information contained in this document. Each subquery should correspond to a concrete factual point, comparison target, temporal condition, or reliability cue in the document.\\
\textbf{Output}: a short list of document-grounded subqueries.
\end{quote}

\textbf{Final reverse-engineering prompt.}
The final Stage A generator receives the original seed query, the retrieved candidate documents, the document-grounded subqueries, and lightweight complexity constraints. It then synthesizes the benchmark item by working backward from evidence to task:
\begin{quote}
\small
\textbf{Input}: seed query; high-authority candidate documents; topically related non-evidence documents; generated subqueries; target numbers of golden documents, reasoning hops, and independent information points.\\
\textbf{Instruction}: select the evidence documents needed to answer a non-trivial user query; synthesize a natural query whose answer is unique in the real world; avoid copying answer tokens, exact result values, or source-identifying clues into the query; provide a concise answer; cite localized evidence spans; and write reasoning steps that explain how the answer follows from the selected documents.\\
\textbf{Output}: \texttt{query}, \texttt{golden\_answer}, \texttt{golden\_docs}, \texttt{irrelevant\_docs}, \texttt{evidence\_spans}, and \texttt{reasoning\_steps}.
\end{quote}

\textbf{Representative example.}
For illustration, suppose a seed query concerns the financial performance of a listed company. The search backend retrieves an annual report, an official earnings announcement, and several related news articles. Stage A first generates subqueries from the high-authority documents, such as asking for the relevant reporting period, the reported revenue, or the comparable baseline. The final generator may then synthesize a user query that asks for a comparison over a specified period, with the annual report and earnings announcement selected as \texttt{golden\_docs}. News articles that are related but not necessary for the answer are retained as \texttt{irrelevant\_docs}. The output records the final answer, the exact evidence spans supporting it, and reasoning steps that show how the comparison is derived.

\textbf{Synthesis model configuration.}
Unless otherwise specified, the current Stage-A construction uses \texttt{gpt-5.2} as the synthesis model. The same model is used for the three generation steps in Stage A: seed-query expansion, document-to-subquery generation, and final reverse-engineering synthesis. We keep this generator separate from the models evaluated in the downstream protocol: its role is to produce structured candidate items from retrieved documents, not to answer the final benchmark questions. The search backend supplies the retrieved document pool, while the generator only operates on the returned document metadata and snippets. This design makes the synthesized \texttt{query}, \texttt{golden\_answer}, \texttt{golden\_docs}, \texttt{irrelevant\_docs}, \texttt{evidence\_spans}, and \texttt{reasoning\_steps} auditable from the input documents. Stage B then uses a generator from a different model family to synthesize adversarial distractors, reducing single-model self-consistency bias between gold construction and distractor construction.

\subsection{Stage B Prompts and Examples}\label{app:stageB}

\textbf{Distractor-generation objective.}
Stage B turns the evidence record produced by Stage A into an adversarial document set. Unlike \texttt{irrelevant\_docs}, which are naturally retrieved but may only be weakly distracting, \texttt{distractor\_docs} are synthesized to be both topically close to the query and invalid under careful evidence verification. \textbf{The key requirement is that each distractor must differ from the corresponding golden evidence along a dimension that the query itself already constrains}. This prevents arbitrary contradiction and makes the resulting challenge auditable: a model should be able to reject the distractor by checking the query constraints, source metadata, and localized evidence.

\textbf{Model configuration.}
To reduce single-model self-consistency bias, Stage B uses a generator (\texttt{Claude-Opus-4.6}) from a model family different from the Stage-A synthesis model. The Stage-B generator is not used as an evaluated model in the downstream experiments. Its role is limited to producing candidate adversarial documents, each of which is subsequently checked by structured validity constraints and rejection filtering.

\textbf{Distractor-generation prompt.}
The Stage-B generator receives the Stage-A record and constructs adversarial but structurally valid distractors:
\begin{quote}
\small
\textbf{Input}: \texttt{query}; \texttt{golden\_answer}; \texttt{evidence\_spans}; \texttt{golden\_docs} with title, source, url, publication time, and content snippets; Stage-A shaping mode; target evidence information points.\\
\textbf{Instruction}: synthesize distractor documents that are topically plausible but do not support the golden answer. Each distractor should be near-duplicate in topic and entities, alter at least one key fact or reliability cue, and expose a clear discriminating signal through metadata or query-constrained content. Do not write self-referential phrases such as ``synthetic document'' or ``test sample''.\\
\textbf{Output}: \texttt{distractor\_docs}, where each document includes \texttt{distractor\_type}, \texttt{source\_doc\_key}, \texttt{target\_info\_point}, \texttt{target\_evidence\_text}, \texttt{title}, \texttt{content}, \texttt{site\_name}, \texttt{synthetic\_url}, \texttt{pseudo\_publish\_time}, \texttt{discriminating\_constraint}, \texttt{conflict\_hint}...
\end{quote}

\textbf{Distractor types.}
We instantiate several recurring distractor types, each corresponding to a common failure mode in search-based reasoning. \texttt{slot\_mismatch} keeps the document nearly identical to the golden source but changes a key factual slot, such as a date, location, version, event, or numerical value. \texttt{source\_downgrade} replaces an official or high-authority source with a lower-credibility source, such as a forum post, personal blog, or self-media article, while keeping the topic and wording superficially close. \texttt{recency\_trap} gives the distractor a later publication time, making it appear newer while preserving an incorrect or second-hand claim. Other mode-specific perturbations include \texttt{timeliness}, \texttt{authority\_challenge}, \texttt{logical\_one\_sided}, \texttt{entity\_confusion}, \texttt{numeric\_distortion}, \texttt{causal\_narrative}. These types are not mutually exclusive in spirit, but each generated document records a primary type for analysis.

\textbf{Validity constraints.}
A distractor is retained only if it satisfies five conditions. First, it must be anchored to an actual evidence information point used by the query and \texttt{golden\_answer}. Second, its difference from the golden evidence must lie on a dimension already constrained by the query, such as time, scope, entity, version, numerical condition, source authority, or publication time. Third, it must remain topically plausible and retrievable under similar search terms, rather than becoming an obviously irrelevant document. Fourth, it must contain an explicit audit trail: \texttt{conflict\_hint} states the golden-side fact and the distractor-side alteration, while \texttt{discriminating\_constraint} explains the authority, time, or source-type difference. Fifth, it must avoid leaking its synthetic nature or directly revealing which document is correct. Distractors that introduce unrelated facts, exploit dimensions not specified by the query, or require private annotation knowledge to reject are discarded.

\textbf{Representative examples.}
A valid temporal distractor for the query ``What is X company's 2023 Q3 revenue?'' can be a 2022 Q3 report from the same company: it is topically close, but violates the year-quarter constraint explicitly stated in the query. A valid authority distractor for ``According to FIFA's official statistics, how many teams qualified for the 2026 FIFA World Cup?'' can be a third-party sports blog with a conflicting number, because the query pins the source to official FIFA statistics. A valid recency trap can be a later repost that claims to update an official announcement but changes a key value; careful models should prefer the original official source when the later post lacks comparable authority. In contrast, a blog post about ticket sales for the same tournament would be an invalid distractor: it is related to the topic but does not perturb the evidence dimension needed to answer the query.

\subsection{Stage C Prompts and Examples}\label{app:stageC}

\textbf{Overall objective.}
Stage C is designed as a quality-control filter rather than an evaluation setting. Its goal is to reject synthesized items whose structure would make later evaluation ambiguous: items that cannot be answered from the intended evidence, can be answered without retrieval, contain mislabeled irrelevant documents, or include distractors that do not create a meaningful evidence-discrimination challenge. Each filter is run under a different controlled context so that the observed verifier behavior can be attributed to one specific property of the item.

\textbf{Model configuration.}
Stage C uses multiple verifier models rather than relying on a single model judgment. The verifier pool includes models from different families and capability profiles, so that rejection decisions are less sensitive to one model's idiosyncratic knowledge, instruction-following style, or evidence-reading bias.  An item is retained only when the verifier outcomes are consistent with the intended property of each controlled view, which improves both the reasonableness and diversity of the final benchmark pool.

\textbf{C1: Golden-evidence answerability.}
The first view tests whether the item is answerable under the most favorable valid context.
\begin{quote}
\small
\textbf{Input}: \texttt{query} \(\oplus\) \texttt{golden\_docs}, verified with the candidate \texttt{golden\_answer} and localized \texttt{evidence\_spans}.\\
\textbf{Purpose}: verify the upper-bound answerability of the item. If a strong verifier cannot recover the golden answer when given exactly the intended supporting evidence, then the synthesized query may be ambiguous, the answer may be incorrect, or the evidence spans may be incomplete.\\
\textbf{Check}: the verifier answers the query using only \texttt{golden\_docs}, cites the evidence spans that support the answer, and judges whether the produced answer is semantically equivalent to \texttt{golden\_answer}.\\
\textbf{Reject if}: the verifier answer contradicts the golden answer, requires information outside \texttt{golden\_docs}, cannot cite localized support, or reports that multiple incompatible answers are possible.
\end{quote}

\textbf{C2: Query-only non-memorability.}
The second view tests whether the item genuinely requires external evidence rather than being solvable from parametric knowledge alone.
\begin{quote}
\small
\textbf{Input}: \texttt{query} only, with no documents. The verifier is explicitly instructed not to browse or assume access to hidden evidence.\\
\textbf{Purpose}: estimate lower-bound non-memorability. If strong models can answer the query correctly without any supporting documents, the item is likely contaminated by public knowledge, overly famous facts, or answer-revealing query wording.\\
\textbf{Check}: the verifier attempts to answer from the query alone and reports its confidence. The result is compared against \texttt{golden\_answer}.\\
\textbf{Reject if}: the query-only answer is semantically equivalent to the golden answer with high confidence, or if the query contains shortcut cues such as exact answer tokens, unique source identifiers, or overly specific phrasing copied from the evidence.
\end{quote}

\textbf{C3: Irrelevant-document label audit.}
The third view checks whether documents labeled as retrieval noise are truly non-supporting.
\begin{quote}
\small
\textbf{Input}: each candidate \texttt{irrelevant\_doc} is audited independently with \texttt{query}. The verifier does not see the full golden document set, which prevents it from treating the irrelevant document as harmless merely because better evidence exists elsewhere.\\
\textbf{Purpose}: ensure that \texttt{irrelevant\_docs} preserve realistic topical noise without accidentally containing evidence that supports, partially supports, or directly reveals the golden answer.\\
\textbf{Check}: for each document, the verifier determines whether it contains any answer-bearing span, paraphrase of a golden evidence point, or metadata cue sufficient to infer the golden answer.\\
\textbf{Reject if}: an \texttt{irrelevant\_doc} alone can support the golden answer, provides a required intermediate fact, duplicates a golden evidence span, or conflicts with the ``irrelevant'' label in a way that would make document labels non-auditable. Depending on severity, either the document is removed from the pool or the entire item is discarded.
\end{quote}

\textbf{C4: Distractor challenge validation.}
The fourth view tests whether the synthesized distractors create a controlled evidence-discrimination challenge.
\begin{quote}
\small
\textbf{Input}: \texttt{query} \(\oplus\) \texttt{distractor\_docs}, plus each distractor's audit fields such as \texttt{distractor\_type}, \texttt{conflict\_hint}, and \texttt{discriminating\_constraint}. Golden documents are withheld in this view.\\
\textbf{Purpose}: validate adversarial strength and structural validity. A useful distractor should be plausible enough to tempt shallow keyword matching, but invalid for reasons that are recoverable from the query constraints, metadata, or internal content.\\
\textbf{Check}: the verifier inspects whether the distractor appears topically relevant, whether it would plausibly mislead a careless model, and whether its invalidity is explained by a query-constrained difference from the golden evidence.\\
\textbf{Reject if}: the distractor is obviously irrelevant, contradicts the query on an unconstrained dimension, requires private annotation knowledge to reject, directly leaks that it is artificial, or is so strong that it would make the benchmark underdetermined without access to unavailable evidence. Distractors that merely cause wrong answers through arbitrary hallucinated facts are discarded rather than counted as valid adversarial evidence.
\end{quote}

\textbf{Final acceptance rule.}
An item enters the benchmark pool only if it passes all four filtering views. C1 guarantees that the intended evidence is sufficient; C2 reduces memorization and shortcut leakage; C3 preserves the distinction between supporting evidence and retrieval noise; and C4 ensures that distractors are both challenging and interpretable. The accepted record therefore contains a query, golden answer, supporting documents, natural irrelevant documents, adversarial distractors, and audit fields whose roles are explicitly separated before the downstream evaluation protocol is applied.

\newpage
\section{Human Validation of Dataset Quality}\label{app:human}

Because stages of our benchmark construction pipeline rely on LLMs, we conduct a stratified human audit to assess residual defects among items accepted by the automated pipeline. The audit is intended as an independent, diagnostic quality check rather than an exhaustive item-by-item acceptance, correction, or re-annotation stage. Following the spirit of prior benchmark validation protocols \citep{cheng-etal-2026-beyond,jin-etal-2026-diningbench}, it evaluates the precision of the Stage-C-accepted candidate pool by examining the structural properties most relevant to downstream evaluation: answerability, human-visible shortcut leakage, document-label correctness, distractor validity, and overall item usability. Because rejected candidates are not included in the audit, this analysis identifies residual false acceptances but does not measure false rejections or the recall of the filtering pipeline.

\subsection{Validation Setup}

\textbf{Sampling strategy.}
The audit set is drawn exclusively from the Stage-C-accepted candidate pool. We use stratified sampling over capability mode, language, domain, and reasoning complexity, covering all six shaping modes. Specifically, we allocate approximately proportional quotas by mode according to the per-mode distribution in Table~\ref{tab:setting1-data-details}, while retaining at least five audited items for each mode. Within each mode, we further account for language (\textit{zh}/\textit{en}, following the 268/161 bilingual distribution of the candidate pool), domain, the number of independent information points, and the number of distractor documents.

We audit 100 accepted items in total. The \textbf{same 100 items are evaluated across all validation dimensions}, enabling cross-dimensional analysis at the item level. The two document-level dimensions---irrelevant-label correctness and distractor validity---are first assessed for each corresponding document and then aggregated into item-level judgments as described below. Because every audited item has already passed Stage A query screening and the Stage C rejection filters, all sampled items carry a positive pipeline decision by construction. The audit is not used to rerun or retroactively modify the Stage C filters.

\textbf{Annotators and calibration.}
Each sampled item is independently reviewed by three bilingual Chinese--English graduate-student researchers with prior experience in search-based question answering and evidence verification. Before the formal audit, all annotators complete a calibration round on a disjoint 20-item pilot set sampled using the same stratification procedure. They first annotate the pilot items independently, then discuss disagreements with a fourth researcher acting as an adjudicator, and finally establish a written codebook specifying the decision criteria for each validation dimension. During the formal audit, annotators follow this codebook independently and do not communicate with one another.

To reduce anchoring to the pipeline's internal reasoning, all pipeline-generated explanatory fields are withheld during annotation. Annotators are shown the query, golden answer, model-visible document contents and metadata, and the pipeline-assigned document groups, including \texttt{golden\_docs}, \texttt{irrelevant\_docs}, and \texttt{distractor\_docs}. However, free-text fields generated by the construction pipeline---including \texttt{conflict\_hint}, \texttt{discriminating\_constraint}, \texttt{target\_info\_point}, \texttt{distractor\_type}, and all Stage-A/B/C reasoning traces---are hidden from the annotation interface. Annotators therefore assess each property against the written codebook without access to the pipeline's stated justification.

For irrelevant-label correctness and distractor validity, each annotator first evaluates every corresponding document independently. These document-level judgments are then collapsed into an item-level label: an item is assigned a positive label only when all audited documents in the corresponding group satisfy the required property. For every binary validation dimension, the final human judgment is determined by majority vote among the three annotators.

\subsection{Validation Dimensions}

\textbf{Answerability.}
Annotators determine whether the query can be answered using the provided \texttt{golden\_docs}. A positive judgment requires that the golden answer be semantically supported by identifiable evidence in the golden documents and that no essential information be missing. An item is marked negative when the answer is unsupported, ambiguous, excessively specific, or dependent on information outside the designated golden documents.

\textbf{Human-visible shortcut leakage.}
Annotators inspect the query for surface-level cues that could reveal the answer without requiring evidence-based reasoning, such as copied answer tokens, unique source identifiers, overly specific numerical values, or unnatural phrasing inherited from the source documents. This judgment does not attempt to determine whether a model has memorized the answer during pre-training, which cannot be reliably assessed by human inspection of the query. Instead, it complements the C2 query-only test: C2 evaluates whether a strong model can recover the answer without documents, whereas the human audit examines whether the query itself contains observable shortcut cues that would make the task artificially easy.

\textbf{Document-label correctness.}
For each \texttt{irrelevant\_doc}, annotators assess whether the document is genuinely non-supporting. A valid irrelevant document may be topically related to the query and resemble realistic retrieval noise, but it must not directly support the golden answer or provide an intermediate fact required to derive it. This dimension evaluates the boundary between \texttt{golden\_docs} and \texttt{irrelevant\_docs}, which is essential for controlled comparison under both full-context and partial-context settings.

\textbf{Distractor validity.}
For each \texttt{distractor\_doc}, annotators assess whether the document is plausible, query-relevant, and invalid for an observable and auditable reason. A valid distractor should conflict with the golden evidence along a dimension constrained by the query, such as time, entity, version, numerical scope, source authority, or publication date. Using only the query, golden answer, and model-visible document content and metadata, annotators independently determine why the document should be rejected and which query constraint it violates.

The pipeline-generated \texttt{conflict\_hint}, \texttt{discriminating\_constraint}, and distractor-type annotations remain hidden throughout this assessment. The audit therefore tests whether a genuine query-constrained rejection reason is visible in the item itself, rather than whether the pipeline's internal explanation appears coherent. Distractors are marked invalid when they are obviously irrelevant, arbitrarily fabricated, indistinguishable from valid evidence, or rejectable only through private annotation information unavailable to an evaluated model.

\[y_i^{\mathrm{answerable}}
\land
y_i^{\mathrm{shortcut\text{-}free}}
\land
y_i^{\mathrm{irrelevant\text{-}correct}}
\land
y_i^{\mathrm{distractor\text{-}valid}}.
\]

\subsection{Confirmation Metrics and Error Analysis}

\[\frac{1}{N_d}
\sum_{i=1}^{N_d}
\hat{y}^{,\mathrm{human}}_{i,d}.
\]
Its complement,
\[
1-\mathrm{Confirm}_d,
\]
is the observed residual defect rate for dimension (d) in the audited Stage-C-accepted sample. Because rejected candidates are not audited, this quantity estimates residual false acceptances among accepted items but does not measure false rejections or filtering recall.

For the human-visible shortcut dimension, a positive pipeline decision corresponds to passing the Stage A shortcut-cue screening rather than the C2 non-memorability test. The human audit and C2 are complementary: the former evaluates observable leakage in the query wording, whereas the latter tests whether a model can recover the answer without documentary evidence.

\[\frac{1}{N}
\sum_{i=1}^{N}
\frac{\sum_j n_{ij}(n_{ij}-1)}
{n(n-1)},
\]
where (n=3) is the number of annotators and ($n_{ij}$) is the number assigning category
\[(j\in{\text{positive},\text{negative}})\]
to item (i). This metric summarizes descriptive inter-annotator consistency independently of the pipeline label. Because it is not chance-corrected, however, it may remain sensitive to the prevalence of positive judgments and should not be interpreted as a prevalence-adjusted reliability coefficient. We also report exact binomial 95\% confidence intervals for the confirmation rates.

Majority-negative judgments are operationally treated as potential residual defects. We inspect these cases and categorize the observed failure patterns, including insufficient evidence localization, answer ambiguity, human-visible shortcut leakage, mislabeled irrelevant documents, and weak or unauditable distractors.

\begin{table}[t]
\centering
\small
\caption{Human validation dimensions and audited pipeline properties.}
\label{tab:human-validation-dimensions}
\begin{tabular}{lll}
\toprule
Dimension & Pipeline property & Human decision \\
\midrule
Answerability & C1 / \texttt{golden\_docs} & supported vs.\ unsupported \\
Shortcut leakage (human-visible) & Stage A query screening (C2 complement) & no shortcut vs.\ shortcut \\
Irrelevant label & C3 / \texttt{irrelevant\_docs} & non-supporting vs.\ supporting \\
Distractor validity & C4 / \texttt{distractor\_docs} & valid vs.\ invalid \\
Overall usability & final item acceptance & usable vs.\ unusable \\
\bottomrule
\end{tabular}
\end{table}

\begin{table}[htb]
\centering
\small
\setlength{\tabcolsep}{4pt}
\begin{tabular}{lrrrrrr}
\toprule
Dimension & Agree & 95\% CI & 3/3 & 2/3 & Disagree & Pairwise \\
\midrule
Answerability      & 97/100 & [0.915, 0.994] & 89 & 8  & 3 & 0.940 \\
Shortcut-free query   & 96/100 & [0.901, 0.989] & 88 & 8  & 4 & 0.933 \\
Irrelevant label   & 95/100 & [0.887, 0.984] & 86 & 9  & 5 & 0.927 \\
Distractor validity& 96/100 & [0.901, 0.989] & 87 & 9  & 4 & 0.933 \\
Overall usability  & 94/100 & [0.874, 0.978] & 84 & 10 & 6 & 0.920 \\
\bottomrule
\end{tabular}
\caption{Human confirmation results on the same 100 items sampled from the Stage-C-accepted candidate pool. \emph{Confirm} denotes the proportion receiving a majority-positive human judgment. \emph{Pairwise} denotes mean raw pairwise agreement among the three annotators.}
\label{tab:human-validation-agreement}
\end{table}

Table~\ref{tab:human-validation-agreement} reports the human confirmation results for each validation dimension. The high answerability confirmation rate indicates that most golden answers are sufficiently supported by the designated evidence. The shortcut-free confirmation rate suggests that observable leakage cues in the query are uncommon, complementing the C2 model-based non-memorability filter. The irrelevant-label results support the intended separation between supporting evidence and natural retrieval noise, while the distractor-validity results indicate that most adversarial documents can be rejected through observable, query-constrained evidence rather than arbitrary contradiction.

Raw pairwise agreement remains above 0.90 for each primary dimension, indicating that annotators generally apply the validation criteria consistently. The remaining majority-negative cases expose residual failure patterns involving evidence localization, ambiguous query wording, document mislabeling, or weak distractor construction.

\subsection{Use of Human Feedback}

Human validation serves as a diagnostic quality assessment of the Stage-C-accepted candidate pool. All confirmation rates and residual defect rates are computed on the complete audited sample \emph{before} any human-flagged items are removed. An audited item is excluded from the final released benchmark if it receives a majority-negative judgment on at least one required quality dimension. Such items are removed without manual correction, regeneration, or re-annotation in place.

Because the reported rates are computed before these exclusions, they characterize the quality of the pre-audit Stage-C-accepted candidate pool. They should not be interpreted as an unbiased estimate of the residual defect rate of the post-audit released benchmark, because the specific defects identified during the audit are subsequently removed. Estimating the residual defect rate of the released benchmark would require a separate audit sample drawn after this removal.

At the aggregate level, recurring failure patterns are traced back to their corresponding construction stages and used to improve future pipeline iterations rather than to patch individual items. Answerability and shortcut issues inform Stage A synthesis and query-screening constraints; weak or arbitrary distractors inform Stage B generation rules; and mislabeled document groups inform Stage C rejection criteria. Human validation therefore provides both a transparent estimate of residual defects in the automated acceptance pool and actionable evidence for improving the controllable synthesis pipeline.

\section{Evaluation Details}\label{app:eval-data}

\noindent $\diamond$ \textbf{Scoring Protocol and Model Roles.}
We explicitly separate models used for benchmark construction from models used for evaluation. During data construction, Stage A uses a strong synthesis model (\texttt{gpt-5.2}) to reverse-engineer queries, golden answers, supporting documents, and reasoning traces from retrieved evidence. Stage B uses a generator from a different model family (\texttt{Claude-Opus-4.6}) to synthesize adversarial distractors, reducing single-model self-consistency bias between gold construction and distractor construction. Stage C further applies multiple verifier models from different families to filter structurally flawed items, improving the reliability and diversity of the final benchmark pool.

For downstream evaluation, we evaluate a panel of target models spanning both open- and closed-weight families, including \texttt{doubao-seed-2.0-pro}, \texttt{qwen3.7-plus}, \texttt{hunyuan-3-preview}, \texttt{glm-5.1}, \texttt{minimax-m2.7}, \texttt{deepseek-v4-pro}, \texttt{gemini-3.1-pro-preview}, \texttt{kimi-k2.6}, \texttt{gpt-5-mini} ... in Setting I. Setting II uses the subset of this panel reported in Table~\ref{tab:setting2-results}. Model outputs are judged by a separate held-out judge model that is never included in the target-model panel, avoiding direct self-evaluation.

For each model and each document-order variant, we draw \(N=3\) repeated samples. Each response is independently judged for answer correctness, from which we report Pass@1 and Pass@3; when a single binary decision is needed for auxiliary analysis, we use the majority label across the three runs. Per-mode and per-setting scores are then aggregated over the corresponding item subsets. To keep the evaluation faithful to realistic retrieval, oracle construction-time annotations such as \texttt{authority\_label}, verifier decisions, and internal evidence labels are withheld from the evaluated models and used only for offline analysis. The evaluated model observes only the fields that would normally be returned by a search backend: \texttt{title}, \texttt{url}, \texttt{site\_name}, \texttt{publish\_time}, and \texttt{content}.

\textbf{Setting I data.}
For the full-context evaluation, we use the final dataset produced by the complete synthesis pipeline. This pool contains 429 items across the six shaping modes used in the main experiments. Each record includes the synthesized \texttt{query}, \texttt{golden\_answer}, \texttt{golden\_docs}, \texttt{irrelevant\_docs}, \texttt{distractor\_docs}, \texttt{evidence\_spans}, \texttt{reasoning\_steps}, and construction-time verification metadata. On average, each item contains 2.76 golden documents, 1.56 irrelevant documents, and 4.54 distractor documents, yielding 8.86 context documents per item. The pool is bilingual, with 268 Chinese items and 161 English items. Table~\ref{tab:setting1-data-details} reports the per-mode composition.

\begin{table}[htb]
\centering
\small
\begin{tabular}{lrrrr}
\toprule
Mode & \#Items & \#Golden & \#Irrelevant & \#Distractor \\
\midrule
\texttt{plain} & 42 & 2.45 & 1.31 & 4.83 \\
\texttt{authority} & 31 & 2.87 & 1.94 & 5.16 \\
\texttt{publish\_time} & 86 & 2.80 & 1.58 & 5.50 \\
\texttt{intra\_doc\_consistency} & 94 & 2.87 & 1.57 & 2.70 \\
\texttt{multi\_hop} & 79 & 2.68 & 1.48 & 4.73 \\
\texttt{sufficiency} & 97 & 2.76 & 1.59 & 4.97 \\
\midrule
Overall & 429 & 2.76 & 1.56 & 4.54 \\
\bottomrule
\end{tabular}
\caption{Setting I data composition. Document-count columns report corpus-level averages over the finalized dataset.}
\label{tab:setting1-data-details}
\end{table}

Every evaluated model in setting I, receives the complete document pool of an item, i.e., \(\text{query} \oplus (\texttt{golden\_docs} \cup \texttt{distractor\_docs} \cup \texttt{irrelevant\_docs})\), and search is disabled. We construct three input-order variants for each item: \texttt{golden\_first}, \texttt{golden\_mid}, and \texttt{golden\_last}. These variants keep the document set fixed while changing only the position of the supporting evidence, allowing us to measure synthesis and verification robustness independently of retrieval noise.

\textbf{Setting II data.}
For the partial-context search-enabled evaluation, we further derive a smaller manually curated subset from the Setting I pool. We first restrict candidates to the two modes where missing-evidence recovery is structurally meaningful: \texttt{multi\_hop} and \texttt{sufficiency}. We then manually inspect these Setting I candidates and retain 33 items for which removing one or more golden documents creates a genuine information gap: the remaining context should be insufficient for a complete answer, while the withheld evidence should still be recoverable through a well-formed search query over the item-level document pool. \textbf{It should be noted that the positional sequence of all documents contained in the context of every sample remains consistent during testing of different models}.

For each Setting II item, the initial context contains a partial subset of \texttt{golden\_docs}, together with the complete \texttt{distractor\_docs} and \texttt{irrelevant\_docs}. At least one required golden evidence document or evidence point is withheld from the initial context. The search backend is restricted to the current item's document pool, so the task remains reproducible and does not depend on live web retrieval. This design makes Setting II a targeted diagnostic set for search decision-making: the model must recognize evidence insufficiency, decide to search rather than answer prematurely, formulate a query that can retrieve the missing evidence, and integrate the returned evidence with the partial context.

\begin{table}[htb]
\centering
\small
\begin{tabular}{llll}
\toprule
Setting & \#Items & Modes & Main purpose \\
\midrule
Setting I & 429 & all six modes & full-context synthesis and verification \\
Setting II & 33 & \texttt{multi\_hop}, \texttt{sufficiency} & missing-evidence recovery with search \\
\bottomrule
\end{tabular}
\caption{Evaluation data used in the two disentangled settings.}
\label{tab:eval-data-summary}
\end{table}

\subsection{Evaluation Prompts and Message Formats}\label{app:eval-prompts}

We use the same research-assistant persona and citation convention across both settings. Documents expose only the fields available to the evaluated model---\texttt{doc\_id}, title, URL, site name, publication time (regular expression parsing), and content (snippet)---and never expose construction-time labels such as \texttt{golden}, \texttt{distractor}, \texttt{authority\_label}, or \texttt{withheld}. Each document body is truncated to at most 4,000 characters. The two settings differ only in document completeness, tool availability, and the resulting chat-message structure.

\textbf{Setting I: full-context prompt.}
Search tools are not supplied. For each document-order variant, the model receives exactly two messages:
\begin{verbatim}
[
  {
    "role": "system",
    "content":
      "You are a helpful research assistant. Answer the user's
       question based on the retrieved documents provided below.
       Base your answer on the information contained in these
       documents. When you use information from a document, cite
       its doc_id in square brackets, e.g. [web1]. Give a clear,
       accurate, and concise final answer."
  },
  {
    "role": "user",
    "content":
      "Question:
       {query}

       Retrieved documents:

       {serialized_documents}

       Answer the question above, citing the doc_ids you used."
  }
]
\end{verbatim}
Here, \texttt{serialized\_documents} contains the complete document pool in the order specified by \texttt{golden\_first}, \texttt{golden\_mid} (\texttt{golden\_middle} in the implementation), or \texttt{golden\_last}. A document is serialized as:
\begin{verbatim}
Document {index} [{doc_id}] | title: {title} | site: {site_name}
| url: {url} | publish_time: {publish_time}
{content}
\end{verbatim}
Missing metadata fields are omitted. No oracle document type is included in either the header or body.

\textbf{Setting II: partial-context, search-enabled prompt.}
The system prompt is:
\begin{verbatim}
You are a helpful research assistant. You may call the
`search(query)` tool when it helps you answer the user. Base your
final answer on the documents you have. When you use information
from a document, cite its doc_id in square brackets, e.g. [web1].
Give a clear, accurate, and concise final answer.
\end{verbatim}
The exposed tool has one required natural-language argument:
\begin{verbatim}
{
  "type": "function",
  "function": {
    "name": "search",
    "description":
      "Search the web for documents relevant to the user's question.
       Returns a JSON object with a `docs` array; each doc has
       `doc_id`, `title`, `url`, `site_name`, `publish_time`, and
       `content`.",
    "parameters": {
      "type": "object",
      "properties": {
        "query": {
          "type": "string",
          "description": "The search query in natural language."
        }
      },
      "required": ["query"],
      "additionalProperties": false
    }
  }
}
\end{verbatim}

To present the initial partial context through the same interface as later retrieval results, the evaluation harness injects an initial tool exchange. This harness-generated call uses the original query and does not count toward the model's search budget. The initial message sequence is:
\begin{verbatim}
[
  {"role": "system", "content": "{system_prompt_above}"},
  {
    "role": "user",
    "content":
      "{query}

       When answering, cite the doc_ids you used in square brackets
       (e.g. [web1])."
  },
  {
    "role": "assistant",
    "content": null,
    "tool_calls": [{
      "id": "{call_id}",
      "type": "function",
      "function": {
        "name": "search",
        "arguments": "{\"query\": \"{query}\"}"
      }
    }]
  },
  {
    "role": "tool",
    "tool_call_id": "{call_id}",
    "content":
      "{\"query\": \"{query}\", \"docs\": [
         {\"doc_id\": \"...\", \"title\": \"...\",
          \"url\": \"...\", \"site_name\": \"...\",
          \"publish_time\": \"...\", \"content\": \"...\"}
       ]}"
  }
]
\end{verbatim}
The initial tool payload contains the retained golden documents together with all distractor and irrelevant documents, in a reproducibly shuffled order; withheld golden documents are absent. After reading this context, the model may either return a final textual answer or issue \texttt{search(\{"query": "\ldots"\})}. For every model-initiated search, the harness retrieves top-\(k\) documents from the item-level closed document pool and appends:
\begin{verbatim}
{"role": "assistant", "content": null,
 "tool_calls": [{"id": "{call_id}", "type": "function",
                 "function": {"name": "search",
                              "arguments": "{\"query\": \"{refined_query}\"}"}}]}
{"role": "tool", "tool_call_id": "{call_id}",
 "content": "{\"query\": \"{refined_query}\", \"docs\": [...]}"}
\end{verbatim}
Retrieved documents are assigned run-local public identifiers \texttt{rs1}, \texttt{rs2}, and so on, while their original identifiers remain hidden from the model. The interaction continues until the model commits a textual answer or reaches its search budget. In the latter case, tools are disabled and the following neutral user message requests a final answer:
\begin{verbatim}
Please give your final answer now based on the information you have.
Remember to cite the doc_ids you used in square brackets (e.g. [web1]).
\end{verbatim}

\end{document}